\documentclass[11pt,letterpaper]{article}
\usepackage{emnlp2016}
\usepackage{times}
\usepackage{latexsym}

\usepackage{graphicx}
\usepackage{hyperref}
\usepackage{amsmath}

\emnlpfinalcopy

\def\emnlppaperid{23}

\title{Gov2Vec: Learning Distributed Representations of Institutions and Their Legal Text\Thanks{Forthcoming paper in the 2016 \emph{Proceedings of Empirical Methods in Natural Language Processing} Workshop on Natural Language Processing and Computational Social Science.}}

\author{John J. Nay \\ 
 School of Engineering \\
  Vanderbilt University \\
 Program on Law \& Innovation\\
Vanderbilt  Law School \\
Nashville, TN 37235, USA\\
 {\tt john.j.nay@gmail.com} \\
 {\tt \href{http://johnjnay.com/}{johnjnay.com}} 
}

\date{}

\begin{document}

\maketitle

\begin{abstract}
We compare policy differences across institutions by embedding representations of the entire legal corpus of each institution and the vocabulary shared across all corpora into a continuous vector space. We apply our method, Gov2Vec, to Supreme Court opinions, Presidential actions, and official summaries of Congressional bills. The model discerns meaningful differences between government branches. We also learn representations for more fine-grained word sources: individual Presidents and (2-year) Congresses. The similarities between learned representations of Congresses over time and sitting Presidents are negatively correlated with the bill veto rate, and the temporal ordering of Presidents and Congresses was implicitly learned from only text. With the resulting vectors we answer questions such as: how does Obama and the 113th House differ in addressing climate change and how does this vary from environmental or economic perspectives? Our work illustrates vector-arithmetic-based investigations of complex relationships between word sources based on their texts. We are extending this to create a more comprehensive legal semantic map.
\end{abstract}

\section{Introduction}\label{introduction}

Methods have been developed to efficiently obtain representations of words in \(\mathbf{R}^d\) that
capture subtle semantics across the dimensions of the vectors (Collobert and Weston, 2008). For instance, after sufficient training, relationships encoded in difference vectors can be uncovered with vector arithmetic: vec(``king'') - vec(``man'') + vec(``woman'') returns a vector close to vec(``queen'') (Mikolov et al.~2013a).

Applying this powerful notion of distributed continuous vector space representations of words, we
embed representations of institutions and the words from their law and policy documents
into shared semantic space. We can then combine positively and
negatively weighted word and government vectors into the same query,
enabling complex, targeted and subtle similarity computations. For instance, which government branch is more characterized by ``validity and truth,'' or ``long-term government career''? 

We apply this method, Gov2Vec, to a unique corpus of Supreme Court opinions, Presidential
actions, and official summaries of Congressional bills. The model discerns meaningful differences between House, Senate, President and Court vectors. We also learn more fine-grained institutional representations: individual Presidents and Congresses (2-year terms). 
The method implicitly learns important latent relationships between these government actors that was not provided during training. For instance, their temporal ordering was learned from only their text. The resulting vectors are used to explore differences between actors with respect to policy topics.

\section{Methods}\label{methods}

A common method for learning vector representations of words is to use a neural network to predict a target word with the mean of its context words' vectors, obtain the gradient with back-propagation of the prediction errors, and update vectors in the direction of higher probability of observing the correct target word (Bengio et al.~2003; Mikolov et al.~2013b). After iterating over many word contexts, words with similar meaning are embedded in similar locations in vector space as a by-product of the prediction task (Mikolov et al.~2013b). Le and Mikolov (2014) extend this word2vec method to learn representations of documents. For predictions of target words, a vector unique to the document is concatenated with context word vectors and subsequently updated. Similarly, we embed institutions and their words into a shared vector space by averaging a vector unique to an institution with context word vectors when predicting that institution's words and, with back-propagation and stochastic gradient descent, update representations for institutions and the words (which are shared across all institutions).\footnote{We use a binary Huffman tree (Mikolov et al.~2013b) for efficient hierarchical softmax prediction of words, and conduct 25 epochs while linearly decreasing the learning rate from 0.025 to 0.001.}

\begin{figure}[t]
\includegraphics[width=3in]{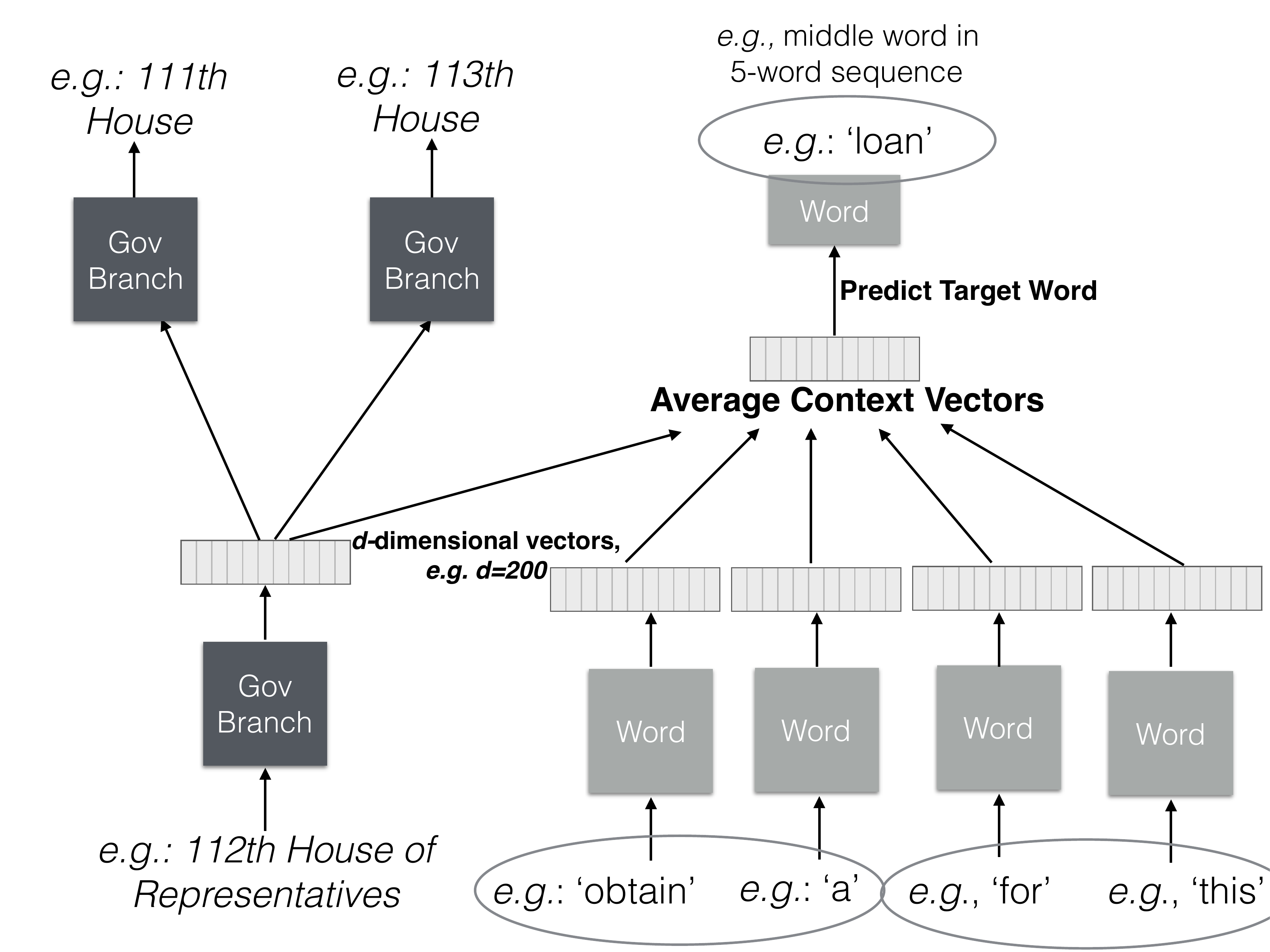}
\caption{Gov2Vec only updates GovVecs with word prediction. For Structured Gov2Vec training, which updates GovVecs with word and Gov prediction, we set ``Gov window size'' to 1, e.g.~a Congress is used to predict those directly before and after.}
\end{figure}

There are two hyper-parameters for the algorithm that can strongly affect results, but suitable values are unknown. 
We use a tree of Parzen estimators search algorithm (Bergstra et al.~2013) to sample from parameter space\footnote{vector dimensionality, uniform(100, 200), and
maximum distance between the context and target words, uniform(10, 25)} and save
all models estimated. Subsequent analyses are conducted across all models, propagating our
uncertainty in hyper-parameters. 
Due to stochasticity in training and the uncertainty in the hyper-parameter values, patterns robust across the ensemble are more likely to reflect useful regularities than individual models.

Gov2Vec can be applied to more fine-grained categories than entire
government branches. In this context, there are often relationships between word
sources, e.g. Obama after Bush, that we can incorporate into the
learning process. During training, we alternate between updating GovVecs
based on their use in the prediction of words in their policy corpus
and their use in the prediction of other word sources located nearby
in time. We model temporal institutional relationships, but any known relationships between entities, e.g.~ranking
Congresses by number of Republicans, could also be incorporated into the
Structured Gov2Vec training process (Fig.~1).

After training, we extract \((M + S) \times d_j \times J\) parameters,
where \(M\) is the number of unique words, \(S\) is the number of word sources, and \(d_j\) the vector
dimensionality, which varies across the \(J\) models (we set $J=20$). We then
investigate the most cosine similar words to particular vector
combinations, \(\arg\max_{v* \in V_{1:N}}{cos(v*, \frac{1}{W} \sum_{i=1}^{W} w_i \times s_i)}\),
where \(cos(a, b) = \frac{\vec{a} \cdot \vec{b}}{\lVert \vec{a}  \rVert \lVert \vec{b}  \rVert}\), \(w_i\) is one of \(W\) WordVecs or GovVecs of interest,
\(V_{1:N}\) are the \(N\) most frequent words in the vocabulary of \(M\)
words (\(N<M\) to exclude rare words during analysis) excluding the \(W\) query words, \(s_i\) is \emph{1} or
\emph{-1} for whether we're positively or negatively weighting \(w_i\).
We repeat similarity queries over all \(J\) models, retain words with $> C$ cosine similarity, and rank the word results
based on their frequency and mean cosine similarity across the ensemble. We also measure the similarity of WordVec combinations to each GovVec and the similarities between GovVecs to validate that the process learns useful embeddings that capture expected relationships.

\section{Data}\label{data}

We created a unique corpus of 59 years of all U.S. Supreme Court opinions (1937-1975, 1991-2010), 227 years of all U.S. Presidential Memorandum, Determinations, and Proclamations, and Executive Orders (1789-2015), and 42 years of official summaries of all bills introduced in the U.S. Congress (1973-2014). We used official summaries rather than full bill text because full texts are only available from 1993 and summaries are available from 1973. We scraped all Presidential Memorandum (1,465), Determinations (801), Executive Orders (5,634), and Proclamations (7,544) from the \href{http://www.presidency.ucsb.edu/}{American Presidency Project website}. The Sunlight Foundation downloaded \href{https://github.com/unitedstates/congress/wiki}{official bill summaries} from the U.S. Government Publishing Office (GPO), which we downloaded. We downloaded Supreme Court Decisions issued 1937--1975 (Vol.~300-422) from the \href{https://www.gpo.gov/fdsys/bulkdata/SCD/1937}{GPO}, and the PDFs of Decisions issued 1991--2010 (Vol.~502-561) from the \href{http://www.supremecourt.gov/opinions/boundvolumes.aspx}{Supreme Court}. We removed HTML artifacts, whitespace, \href{http://jmlr.csail.mit.edu/papers/volume5/lewis04a/a11-smart-stop-list/english.stop}{stop words}, words occurring only once, numbers, and punctuation, and converted to lower-case.

\section{Results}\label{results}

\subsection{WordVec-GovVec Similarities}\label{wordvec-govvec-similarities}

We tested whether our learned vectors captured meaningful differences between branches. Fig.~2 displays similarities between these queries and the branches, which reflect \textit{a priori} known differences. 

Gov2Vec has unique capabilities that summary statistics,
e.g.~word frequency, lack: it can compute similarities between any
source and word as long as the word occurs at least in one
source, whereas word counting cannot provide meaningful similarities when a word never occurs in a source's corpus. Most importantly, Gov2Vec can combine complex combinations of positively and negatively weighted vectors in a similarity query.

\begin{figure}[t]
\includegraphics[width=3.1in]{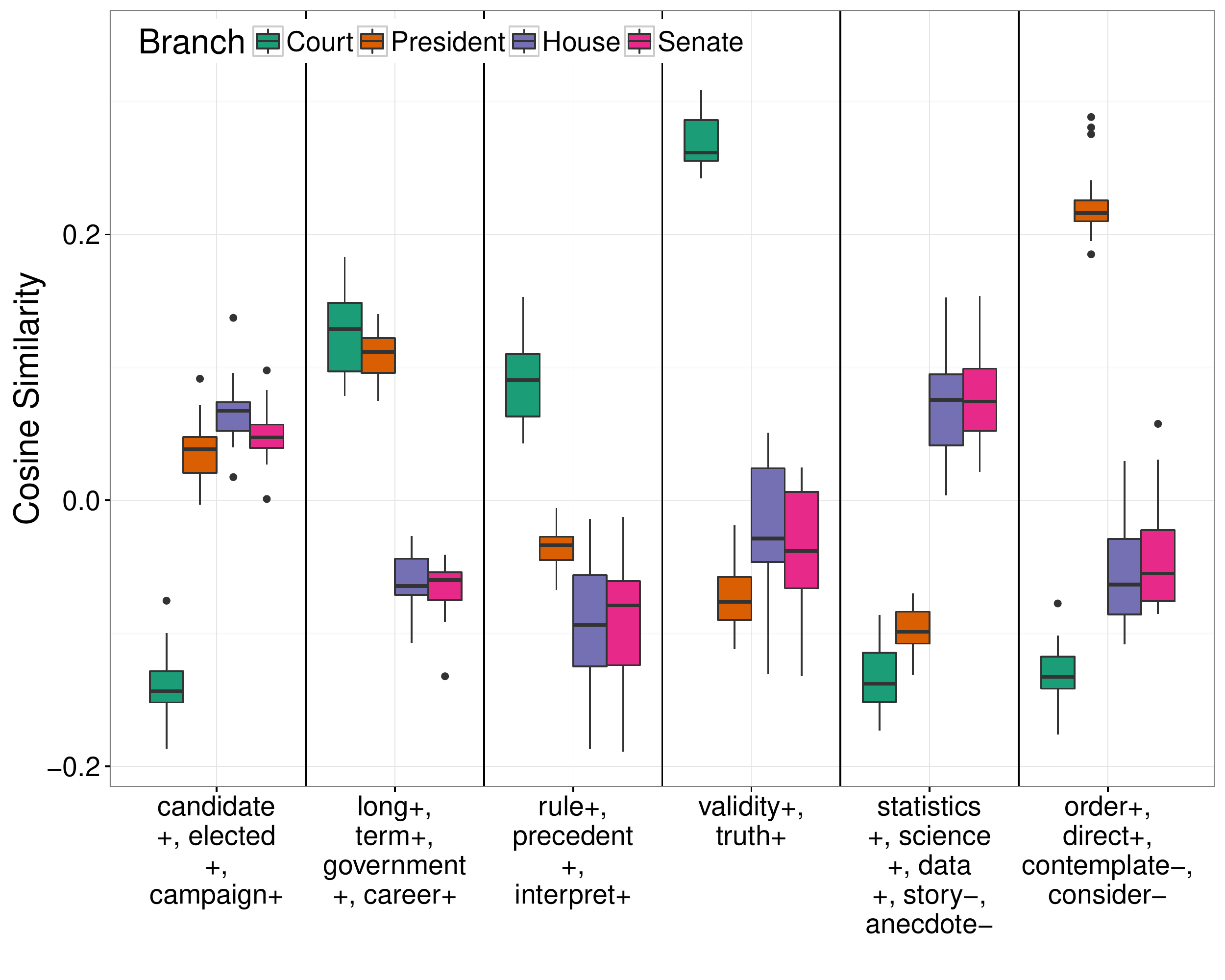}
\caption{We compute the mean of similarities of each branch to all words and subtract this from each branch's similarity computation to normalize branches within vector space. Positive and negative weighting is noted with \texttt{+} and \texttt{-}. Compared to the Court, the President is much closer to ``order and direct'' than ``contemplate and consider.'' The opposite holds for ``validity and truth.'' The left panel reflects the fact that the House is elected most often and the Court the least (never).}
\end{figure}

\subsection{GovVec-GovVec Similarities}\label{govvec-govvec-similarities}

We learned representations for individual Presidents and Congresses by
using vectors for these higher resolution word sources in the
word prediction task. To investigate if the representations capture
important latent relationships between institutions, we compared the
cosine similarities between the Congresses over time (93rd--113th) and
the corresponding sitting Presidents (Nixon--Obama) to the bill
veto rate. We expected that a lower veto rate would be reflected in more similar vectors, and, indeed, the Congress-President similarity and veto rate are negatively correlated (Spearman's \(\rho\) computed on raw veto rates and similarities: -0.74; see also Fig.~3).\footnote{Leveraging temporal relationships in the learning process, Structured Gov2Vec, and just using the text, yield very similar (impressive) results on this task. Figs.~3 and 4 and the correlation reported are derived from the text-only Gov2Vec results.}
 
 \begin{figure}[t]
  \includegraphics[width=2.5in]{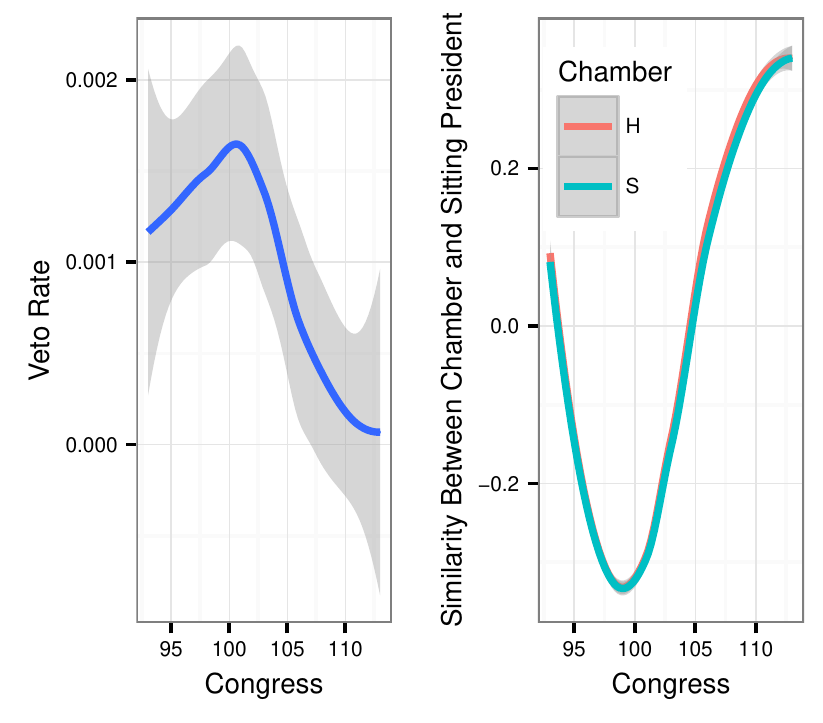}
  \caption{Loess-smoothed veto rate (left) and loess-smoothed (across ensemble) similarity of Congresses to Presidents (right).}
 \end{figure}
 
 \begin{figure}[t]
  \includegraphics[width=3in]{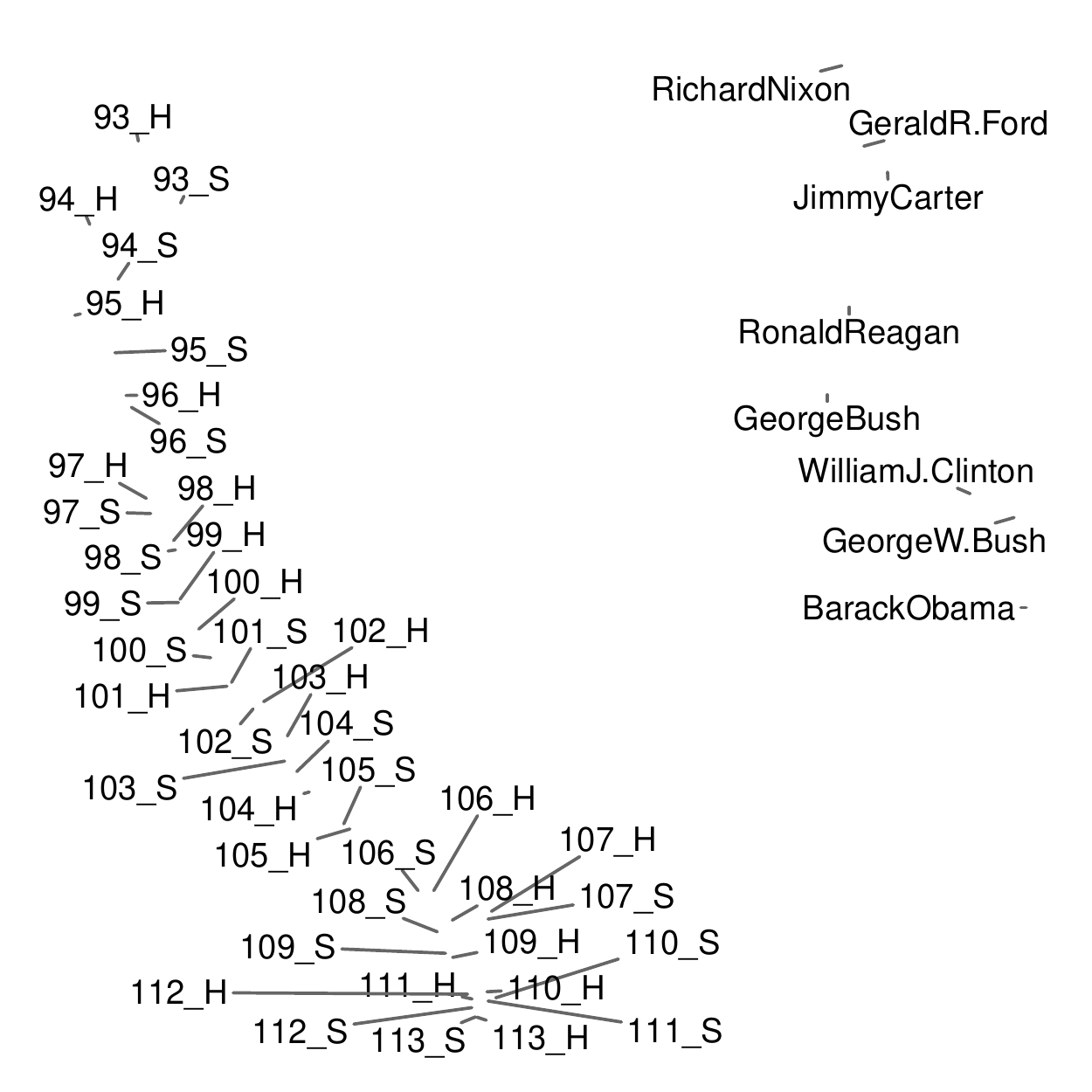}
  \caption{2-dimensional Congresses (93-113 House and Senate) and Presidents.}
\end{figure}

As a third validation, we learn vectors from only text and project them into two dimensions with principal components analysis. From Fig.~4 it's evident that temporal and institutional relationships were implicitly learned.\footnote{These are the only results reported in this paper from a single model within the ensemble. We conducted PCA on other models in the ensemble and the same relationships hold.} One dimension (top-to-bottom) almost perfectly rank orders Presidents and Congresses by time, and another dimension (side-to-side) separates the President from Congress.

\subsection{GovVec-WordVec Policy Queries}\label{complex-govvec-wordvec-queries}

Fig.~5 (top) asks: how does Obama and the 113th House differ in addressing climate change and how does this vary across environmental and economic contexts? The most frequent word across the ensemble (out of words with $>0.35$ similarity to the query) for the Obama-economic quadrant is ``unprecedented.'' ``Greenhouse'' and ``ghg'' are more frequent across models and have a higher mean similarity for Obama-Environmental than 113th House-Environmental. 

Fig.~5 (bottom) asks: how does the House address war from ``oil'' and ``terror'' perspectives and how does this change after the 2001 terrorist attack.\footnote{For comparisons across branches, e.g.~113th House and Obama, Structured Gov2Vec learned qualitatively more useful representations so we plot that here. For comparisons within Branch, e.g.~106th and 107th House, to maximize uniqueness of the word sources to obtain more discriminating words, we use text-only Gov2Vec.} Compared to the 106th, both the oil and terrorism panels in the 107th (when 9-11 occurred) have words more cosine similar to the query (further to the right) suggesting that the 107th House was closer to the topic of war, and the content changes to primarily strong verbs such as instructs, directs, requires, urges, and empowers.

\begin{figure}[t]
\includegraphics[width=3.3in]{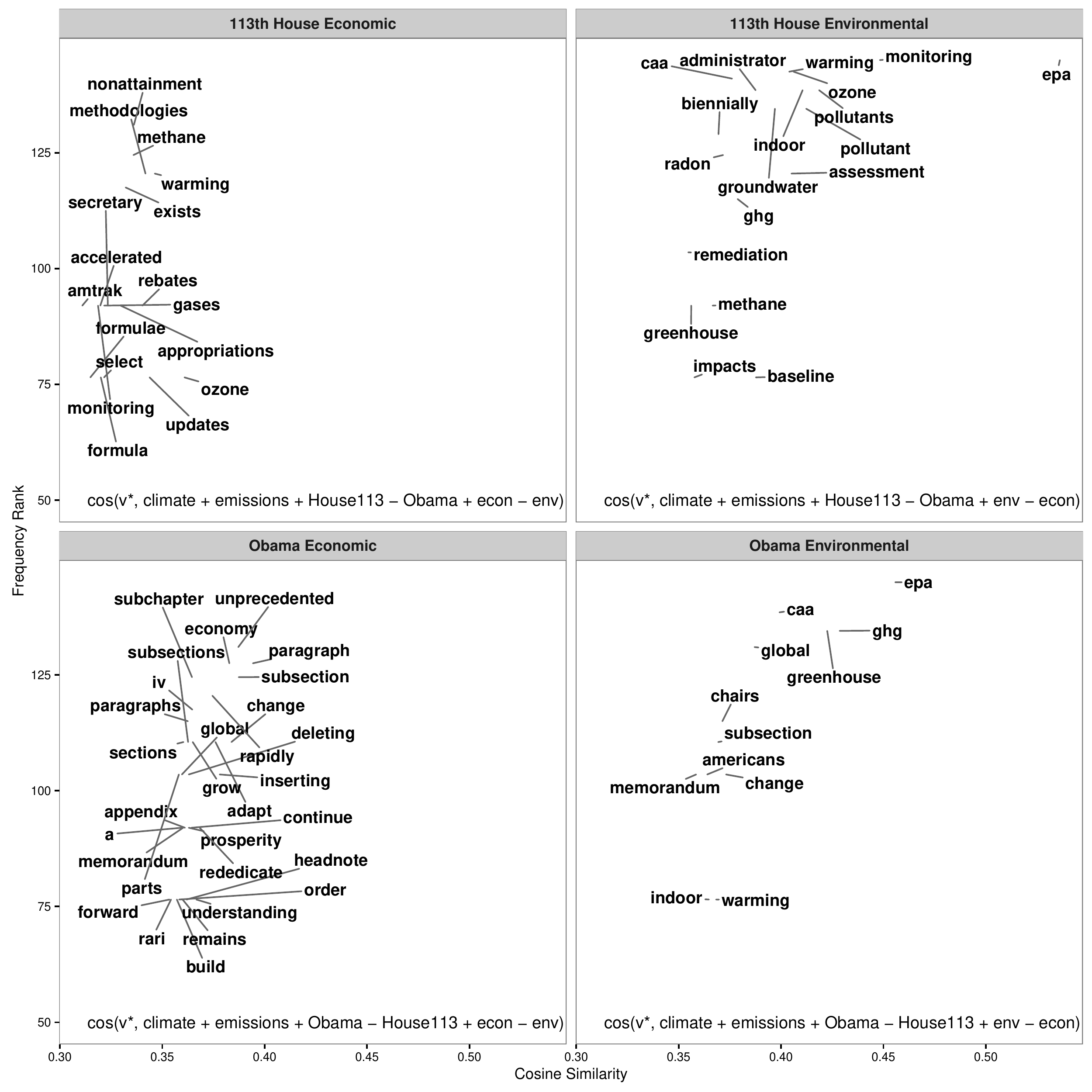}
\includegraphics[width=3.3in]{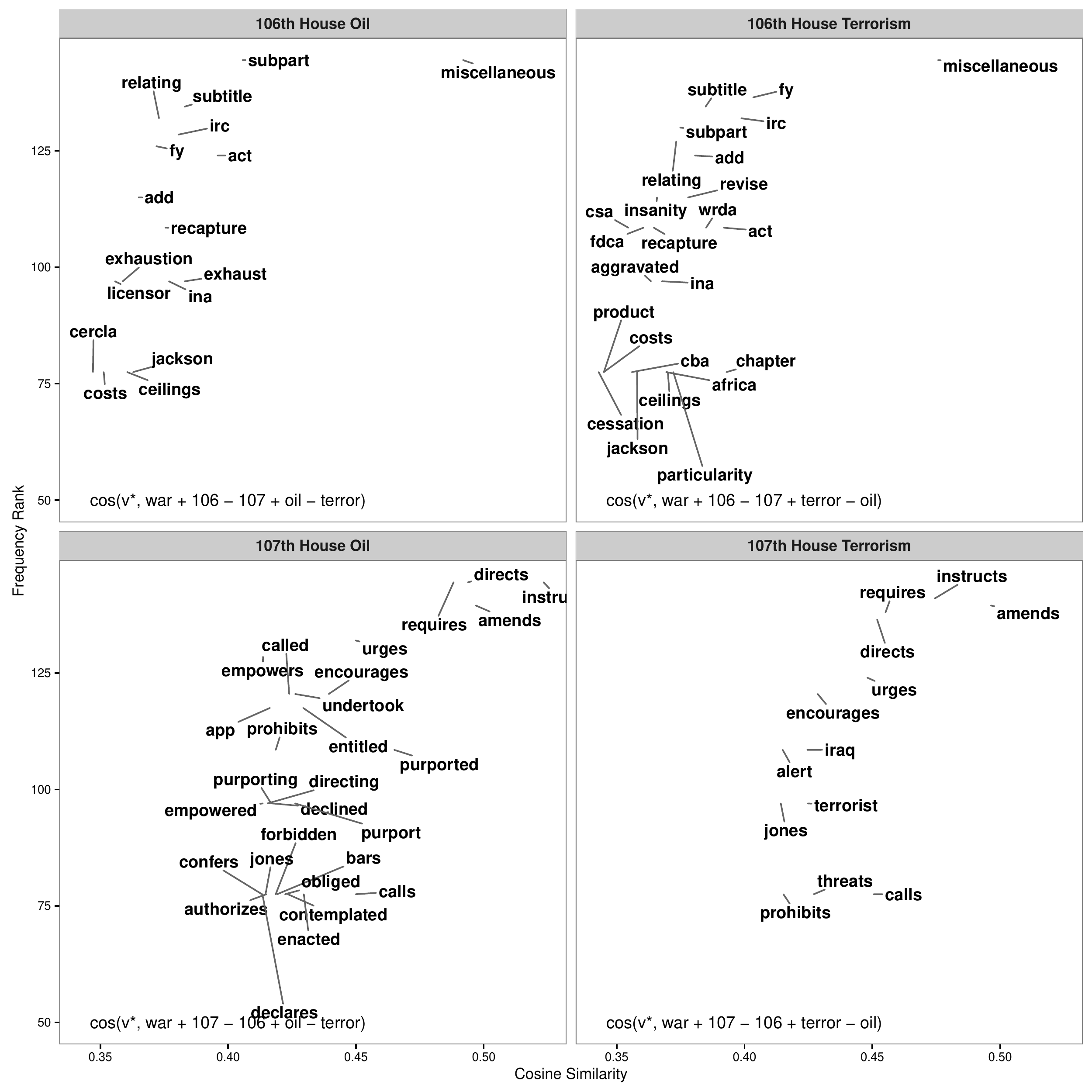}
\caption{The top panel is the climate policy query comparing the 113th U.S. House of Representatives and President Obama: ~\(\arg\max_{v*}\) \(cos(v*, wv(\text{climate}) + wv(\text{emissions}) + gv(\text{G1}) - gv(\text{G2}) + wv(\text{C1}) - wv(\text{C2})\), where \emph{G}=\(\{\text{113 House, Obama}\}\), \(G1 \in G\), \(G2 \in G\), \(G1 \neq G2\), \emph{C}=\(\{\text{economic, environmental}\}\), \(C1 \in C\), \(C2 \in C\), \(C1 \neq C2\). The bottom panel is the war policy query for the U.S. House of Representatives before and after the 9-11 terrorist attacks: wv(\text{war}), \emph{G}=\(\{\text{106 House, 107 House}\}\), \emph{C}=\(\{\text{oil, terror}\}\). The exact query used to create each quadrant is provided at the bottom of the quadrant.}
\end{figure}

\section{Additional Related Work}\label{related-work}

Political scientists model text to understand political processes (Grimmer 2010; Roberts et al. 2014); however, most of this work focuses on variants of topic models (Blei et al. 2003). 
Djuric et al.~(2015) apply a learning procedure similar to Structured
Gov2Vec to streaming documents to learn representations of documents
that are similar to those nearby in time. Structured Gov2Vec applies this
joint hierarchical learning process (using entities to predict words \emph{and}
other entities) to non-textual entities. Kim et al.~(2014) and Kulkarni
et al.~(2015) train neural language models for each year of a time
ordered corpora to detect changes in words. Instead of learning models for
distinct times, we learn a global model with embeddings for
time-dependent entities that can be included in queries to
analyze change. Kiros et al.~(2014) learn embeddings for text attributes by treating them as gating units to a word embedding tensor.  Their process is more computationally intensive than ours.

\section{Conclusions and Future Work}\label{conclusions-and-future-work}

We learned vector representations of text meta-data on a novel data set of legal texts that includes case, statutory, and administrative law. The representations effectively encoded important relationships between institutional actors that were not explicitly provided during training. Finally, we demonstrated fine-grained investigations of policy differences between actors based on vector arithmetic. More generally, the method can be applied to measuring similarity between any entities producing text, and used for recommendations, e.g.~what's the closest \emph{think-tank vector} to the \emph{non-profit vector} representation of the Sierra Club?

Methodologically, our next goal is to explore where training on non-textual relations, i.e. Structural Gov2Vec, is beneficial. It seems to help stabilize representations when exploiting temporal relations, but political relations may prove to be even more useful. Substantively, our goal is to learn a large collection of vectors representing government actors at different resolutions and within different contexts\footnote{For instance, learn a vector for the 111th House using its text and temporal relationships to other Houses, learn a vector for the 111th House using its text and political composition relationships to other Houses (e.g.~ranking by number of Republicans), and then move down in temporal and institutional resolution, e.g. to individual Members. Then use press release text to gain a different perspective and iterate through the resolutions again.}  to address a range of targeted policy queries. Once we learn these representations, researchers could efficiently search for differences in law and policy across time, government branch, and political party.

\section*{Acknowledgments}

We thank the anonymous reviewers for helpful suggestions.

\section*{References}\label{references}
\addcontentsline{toc}{section}{References}

\hypertarget{refs}{}
\hypertarget{ref-bengioux5fneuralux5f2003}{}
Bengio, Yoshua, Réjean Ducharme, Pascal Vincent, and Christian Janvin.
2003. A Neural Probabilistic Language Model. \emph{J. Mach. Learn.
Res.} 3 (March): 1137--55.

\hypertarget{ref-bergstraux5fmakingux5f2013}{}
Bergstra, James S., Daniel Yamins, and David Cox. 2013. Making a
Science of Model Search: Hyperparameter Optimization in Hundreds of
Dimensions for Vision Architectures. In \emph{Proceedings of the 30th
International Conference on Machine Learning}, 115--23.

\hypertarget{ref-bleiux5flatentux5f2003}{} Blei, David M., Andrew Y. Ng, and Michael I. Jordan. 2003. ``Latent Dirichlet Allocation.'' \emph{J. Mach. Learn. Res.} 3 (March): 993--1022.

Collobert, Ronan and Jason Weston. 2008. A Unified Architecture for Natural Language Processing: Deep Neural Networks with Multitask Learning. In \emph{Proceedings of the 25th International Conference on Machine Learning}. 160--167. ACM.

\hypertarget{ref-djuricux5fhierarchicalux5f2015}{}
Djuric, Nemanja, Hao Wu, Vladan Radosavljevic, Mihajlo Grbovic, and
Narayan Bhamidipati. 2015. Hierarchical Neural Language Models for
Joint Representation of Streaming Documents and Their Content. In
\emph{Proceedings of the 24th International Conference on World Wide
Web}, 248--55. WWW '15. New York, NY, USA: ACM.

Grimmer, Justin. 2010. ``A Bayesian Hierarchical Topic Model for Political Texts: Measuring Expressed Agendas in Senate Press Releases.'' \emph{Political Analysis} 18 (1): 1--35.


\hypertarget{ref-kimux5ftemporalux5f2014}{}
Kim, Yoon, Yi-I. Chiu, Kentaro Hanaki, Darshan Hegde, and Slav Petrov.
2014. Temporal Analysis of Language Through Neural Language Models.
In \emph{Proceedings of the ACL 2014 Workshop on Language Technologies
and Computational Social Science}, 61--65. Association for Computational
Linguistics.

\hypertarget{ref-kirosux5fmultiplicativeux5f2014}{}
Kiros, Ryan, Richard Zemel, and Ruslan R Salakhutdinov. 2014. A
Multiplicative Model for Learning Distributed Text-Based Attribute
Representations. In \emph{Advances in Neural Information Processing
Systems 27}, edited by Z. Ghahramani, M. Welling, C. Cortes, N. D.
Lawrence, and K. Q. Weinberger, 2348--56. Curran Associates, Inc.

\hypertarget{ref-kulkarniux5fstatisticallyux5f2015}{}
Kulkarni, Vivek, Rami Al-Rfou, Bryan Perozzi, and Steven Skiena. 2015.
Statistically Significant Detection of Linguistic Change. In
\emph{Proceedings of the 24th International Conference on World Wide
Web}, 625--35. WWW '15. New York, NY, USA: ACM.

\hypertarget{ref-leux5fdistributedux5f2014}{}
Le, Quoc, and Tomas Mikolov. 2014. Distributed Representations of
Sentences and Documents. In \emph{Proceedings of the 31st
International Conference on Machine Learning}, 1188--96.


\hypertarget{ref-mikolovux5flinguisticux5f2013}{}
Mikolov, T., W.T. Yih, and G. Zweig. 2013a. Linguistic Regularities in
Continuous Space Word Representations. In \emph{HLT-NAACL}, 746--51.

\hypertarget{ref-mikolovux5fdistributedux5f2013}{}
Mikolov, Tomas, Ilya Sutskever, Kai Chen, Greg S Corrado, and Jeff Dean.
2013b. Distributed Representations of Words and Phrases and Their
Compositionality. In \emph{Advances in Neural Information Processing
Systems 26}, edited by C. J. C. Burges, L. Bottou, M. Welling, Z.
Ghahramani, and K. Q. Weinberger, 3111--9. Curran Associates, Inc.

\hypertarget{ref-robertsux5fstructuralux5f2014}{} Roberts, Margaret E., Brandon M. Stewart, Dustin Tingley, Christopher Lucas, Jetson Leder-Luis, Shana Kushner Gadarian, Bethany Albertson, and David G. Rand. 2014. ``Structural Topic Models for Open-Ended Survey Responses.'' \emph{American Journal of Political Science} 58 (4): 1064--82.

\end{document}